\documentclass[sigconf, nonacm]{acmart}

\newcommand\vldbdoi{XX.XX/XXX.XX}
\newcommand\vldbpages{XXX-XXX}
\newcommand\vldbvolume{14}
\newcommand\vldbissue{1}
\newcommand\vldbyear{2020}
\newcommand\vldbauthors{\authors}
\newcommand\vldbtitle{\shorttitle} 
\newcommand\vldbavailabilityurl{}
\newcommand\vldbpagestyle{plain} 

\usepackage{listings}
\usepackage{xcolor}
\usepackage{graphicx}
\usepackage{float}

\usepackage{booktabs}
\usepackage{multirow}
\usepackage{pifont}

\lstdefinestyle{codeStyle}{
  basicstyle=\footnotesize\ttfamily,
  numbers=left,
  numberstyle=\tiny,
  numbersep=4pt,
  breaklines=true,
  frame=single,
  backgroundcolor=\color{gray!10},
  captionpos=b
}

\begin{document}
\title{GRAIL: AI translation for scientists application workflow on satellite data}
\author{Zhuocheng Shang}
\affiliation{%
  \institution{University of California, Riverside}
  \city{Riverside}
  \state{California}
}
\email{zshan011@ucr.edu}

\author{Ahmed Eldawy}
\affiliation{%
  \institution{University of California, Riverside}
  \city{Riverside}
  \state{California}
}
\email{eldawy@ucr.edu}

\begin{abstract}
Domain scientists increasingly develop Python scripts to analyze satellite imagery but they lack scalability to large-scale data. This paper demonstrates GRAIL, an agentic translation system that converts Python geospatial workflows into executable Spark-based programs without requiring scientists to learn a new framework. Rather than fine-tuning a specialized LLM model, GRAIL adapts RDPro, a Scala library for satellite data analysis, to make it LLM-ready using structured documentation, API alias functions, and repair-oriented error logs. Translation is structured as a LangGraph pipeline that decomposes code generation into explicit sections with guided inputs and outputs, enabling targeted repair without regenerating the full program. We demonstrate GRAIL on real-world geospatial workflows and showcase the correctness and scalability of the translated code.
\end{abstract}

\maketitle

\pagestyle{\vldbpagestyle}
\begingroup\small\noindent\raggedright\textbf{PVLDB Reference Format:}\\
\vldbauthors. \vldbtitle. PVLDB, \vldbvolume(\vldbissue): \vldbpages, \vldbyear.\\
\href{https://doi.org/\vldbdoi}{doi:\vldbdoi}
\endgroup
\begingroup
\renewcommand\thefootnote{}\footnote{\noindent
This work is licensed under the Creative Commons BY-NC-ND 4.0 International License. Visit \url{https://creativecommons.org/licenses/by-nc-nd/4.0/} to view a copy of this license. For any use beyond those covered by this license, obtain permission by emailing \href{mailto:info@vldb.org}{info@vldb.org}. Copyright is held by the owner/author(s). Publication rights licensed to the VLDB Endowment. \\
\raggedright Proceedings of the VLDB Endowment, Vol. \vldbvolume, No. \vldbissue\ %
ISSN 2150-8097. \\
\href{https://doi.org/\vldbdoi}{doi:\vldbdoi} \\
}\addtocounter{footnote}{-1}\endgroup

\ifdefempty{\vldbavailabilityurl}{}{
\vspace{.3cm}
\begingroup\small\noindent\raggedright\textbf{PVLDB Artifact Availability:}\\
The source code, data, and/or other artifacts have been made available at \url{\vldbavailabilityurl}.
\endgroup
}

\sloppy

\section{Introduction}

Over the past decade, satellite data from missions such as Landsat, Sentinel, and MODIS has grown rapidly and become widely available for research on agriculture, land use, climate change, and disasters. However, the utilization of this data is limited by its massive volume. Most domain scientists use Python scripts that employ single-machine libraries, e.g., Rasterio and GeoPandas, which are great for prototyping and experimentation but do not scale to large parallel computation.


Recent systems such as Apache Sedona~\cite{yu2016demonstration}, GeoTrellis, and RDPro~\cite{shang2024rdpro} support scalable raster processing, but they require users to learn new frameworks, distributed computing concepts, and often Scala or Java, making them difficult for many data scientists to adopt. Recently, we worked closely with agronomists and hydrologists to port a complex model that predicts evapotranspiration (ET) maps to Apache Spark to support large-scale data~\cite{FieldSAT}. However, this approach is difficult to replicate due to the steep learning curve associated with Spark. Another approach is to develop a Python interface that wraps existing scalable functionality in RDPro but this requires substantial ongoing engineering effort and still requires additional learning.


Recent advances in LLMs offer an alternative path forward for making specialized libraries accessible to non-technical users. LLMs can now generate end-to-end programs from simple text descriptions~\cite{trummer2025generating, guo2024deepseek}, motivating this demo to explore translating data scientists' Python scripts, or even plain-text descriptions, directly into Scala code that runs on Spark~\cite{jin2025llms}. However, this approach presents three main challenges that are explored in this demo.
First, LLMs are well trained on common geospatial Python libraries such as GeoPandas and Rasterio, which have tons of examples used in training, but are less familiar with newer research libraries.
Second, code comments and README files are traditionally written for human consumption and may not provide the kind of information LLMs need~\cite{wijaya2025readme}.
Third, while much prior work studies code generation from text, less attention has been given to translating existing code across both programming languages and library ecosystems.



This paper presents GRAIL, a system that we developed to demonstrate and study how LLMs can be used with newly developed libraries that are still under development and do not have the advantage of being used in training.
Rather than focusing on the LLM side, we study how to make the library itself, RDPro~\cite{shang2024rdpro} in our case, LLM-ready so that it can work with existing agents and code generators.
GRAIL builds an agentic pipeline using LangGraph that is tailored for satellite data analysis.
It starts with analyzing the task in-hand, whether a Python script or a textual description, and then follows with a planning phase that uses a scaffold that guides the planning phase. After that, it generates the code section-by-section and finalizes with a validation and repair phase to produce the final code.

In addition to the agent-based code generation, we also extend the RDPro library itself with three new ideas and test their effect on the correctness and quality of code generation.
First, we develop specialized documentation that is targeted towards LLMs rather than humans. They add more details and instructions that are specifically useful for LLM-based code generation.
Second, we extended the code itself with \emph{alias functions} that provide multiple entry points to the same functionality but with common standard terminology that the LLM is familiar with, e.g., using function names and signatures popular with other raster-based libraries.
Finally, we extend error messages to be more verbose and enrich them with instructions on how to fix the error. These detailed error instructions are designed so that a code generator can use as feedback to fix the code on the next iteration.

\section{GRAIL Translation Pipeline}
\label{sec:system}

\begin{figure}[t]
    \centering
    \includegraphics[width=\columnwidth]{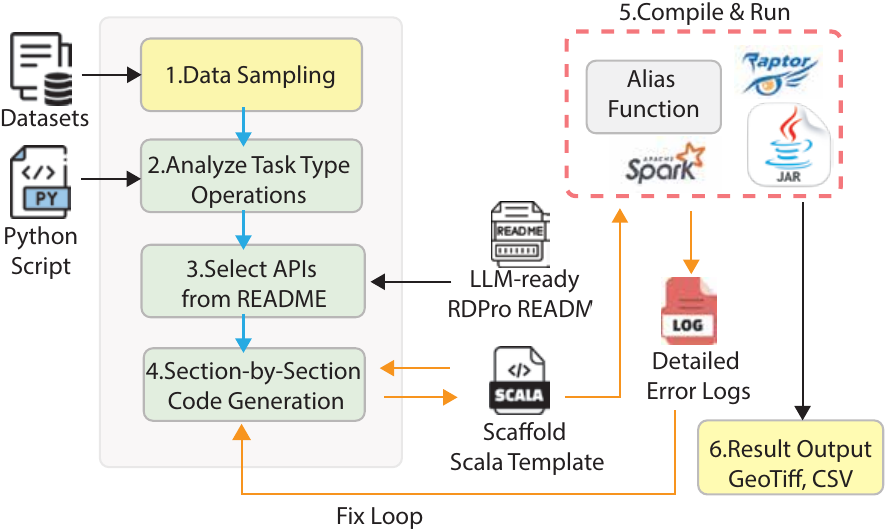}
    \caption{GRAIL translation system overview.}
    \Description{
    Flowchart titled ``GRAIL translation system overview.'' Inputs on the left are datasets and a Python script. The main pipeline has four steps: (1) data sampling, (2) analyze task type and operations, (3) select APIs from README, and (4) section-by-section code generation. An ``LLM-ready RDPro README'' feeds into API selection, and a ``scaffold Scala template'' supports code generation. The generated Scala code is then sent to step (5) compile and run, which includes alias functions and the Raptor, Spark, and Java/JAR environments. Compilation produces detailed error logs, which feed back into code generation through a fix loop. Successful execution produces step (6) result output, shown as GeoTIFF and CSV. Arrows indicate the end-to-end workflow and iterative debugging loop.
    }
    \label{fig:grail}
\end{figure}
Figure~\ref{fig:grail} illustrates the GRAIL pipeline. The system accepts either a Python script or a plain-text task description and produces an executable RDPro Scala program. GRAIL decomposes translation into four stages: task analysis, section planning, section-by-section generation, and iterative repair.

\textbf{Task Analysis.}
GRAIL converts the input, either Python script or plain-text, into a structured intermediate representation that captures the task type and required geospatial operations. Translating the code into this representation reduces the risk of mis-translating source-language function calls directly into the target library.

\textbf{Section Planning.}
A planner analyzes the task and selects relevant sections from the LLM-ready documentation, then instantiates them using a \emph{scaffold file} that defines the general structure of a geospatial job. The scaffold covers six logical stages: \textsc{LoadData} specifies raster and vector inputs; \textsc{TypeCheck} validates pixel types to prevent casting errors; \textsc{SpatialCheck} ensures all datasets share a consistent geospatial extent; \textsc{Transform} handles raster-only preparation; \textsc{RasterVectorJoin} materializes the join between raster and vector data; and \textsc{Analytics} computes the final result, whether a histogram over raster data or an aggregated statistic from the join. Sections irrelevant to the inferred workflow are pruned before generation begins.

\textbf{Section-by-Section Generation.}
Each section is generated sequentially within a LangGraph loop. The prompt for each section incorporates the relevant API documentation fragments retrieved for that section, along with a section contract that specifies required API calls, input variables, expected output formats, and forbidden patterns. It also includes the current scaffold state, including variable summaries from preceding sections. This shared context allows later sections to build upon results generated in earlier sections.

\textbf{Validation and Repair.}
After each candidate section is generated, the pipeline applies four validation layers covering a scope check, a deterministic contract check, an LLM semantic review, and full Spark compilation and submission. If any layer fails, targeted repair feedback is assembled from the detected issues and injected into the next prompt, retrying only the failing section up to five rounds. Once all sections pass, a final whole-program review checks cross section invariants before writing the completed Scala program to disk.
\section{LLM-Ready Library Modifications}
\label{sec:mod}
To improve the reliability of LLM-generated RDPro code, we introduce and demonstrate three complementary modifications: structured documentation, API alias functions, and detailed error logs.

\textbf{Structured Documentation.}
The original RDPro documentation was written for human readers, with extensive descriptions and loosely organized examples. While readable, this format makes it difficult for an LLM to identify function signatures, input constraints, and expected outputs. We redesign the documentation into a structured and consistent format, where each API entry includes the function name, a description, input parameters with types and constraints, the output type and semantics, and a minimal executable usage example. During both planning and code generation, the pipeline carries relevant API fragments from this documentation and injects them directly into the prompt, ensuring the model reasons over precise specifications rather than general assumptions. As shown in \autoref{fig:grail}, the pipeline explicitly draws from the LLM-ready RDPro README at the API selection stage.

\textbf{API Alias Functions.}
During our preliminary experiments, we observed that LLMs tend to directly transfer function usage patterns from well-known Python geospatial libraries, such as GDAL and Rasterio. For example, LLMs loading raster data commonly generate names such as \texttt{open} or \texttt{read}, and use \texttt{mask} for raster-vector join operations, following conventions from Python ecosystems they have seen during training. To bridge this gap, we introduce \emph{alias functions} directly in the RDPro library that map these commonly inferred names to their correct RDPro equivalents, so that generated code executes correctly without requiring a repair round. The example below shows a typical case where a Python-style name is 
transparently redirected to the correct RDPro call, as part of the 
compile and run stage illustrated in \autoref{fig:grail}.



\begin{lstlisting}[style=codeStyle]
// LLM-generated: Rasterio-style masking
val masked = raster.mask(vector)
// Correct RDPro join via raptorJoin
val joinedRecords = raster.raptorJoin(vector)
\end{lstlisting}


\textbf{Repair-Oriented Error Reporting.}
Since LLM-generated code cannot be guaranteed to compile on the first attempt, the quality of repair feedback directly affects how quickly the pipeline converges. Generic runtime errors provide limited guidance for automated correction. For example, they may indicate where an error occurs, but often provide little insight into how to fix it, especially when the issue involves customized internal functions. To address this, we augment selected RDPro error messages with actionable repair hints that indicate both the cause of the failure and a suggested fix. For example, a type mismatch produces:

\begin{lstlisting}[style=codeStyle]
Type mismatch: expected Float but found Int.
Suggested fix: val raster: RasterRDD[Int] = sc.geoTiff()
\end{lstlisting}

When a section fails in compilation or execution, the agent extracts the relevant error, attaches the corresponding repair hint, and injects the combined feedback into the next generation prompt as a repair context block. \autoref{fig:grail} shows how compilation and execution errors 
are captured in structured logs and fed back into the fix loop for 
targeted repair. By providing this targeted signal together with the section contract and re-retrieved API documentation, the model receives both diagnosis and corrective guidance without the need to reason over the full error trace.
\begin{figure*}[!t]
    \centering
    \includegraphics[width=0.85\textwidth]{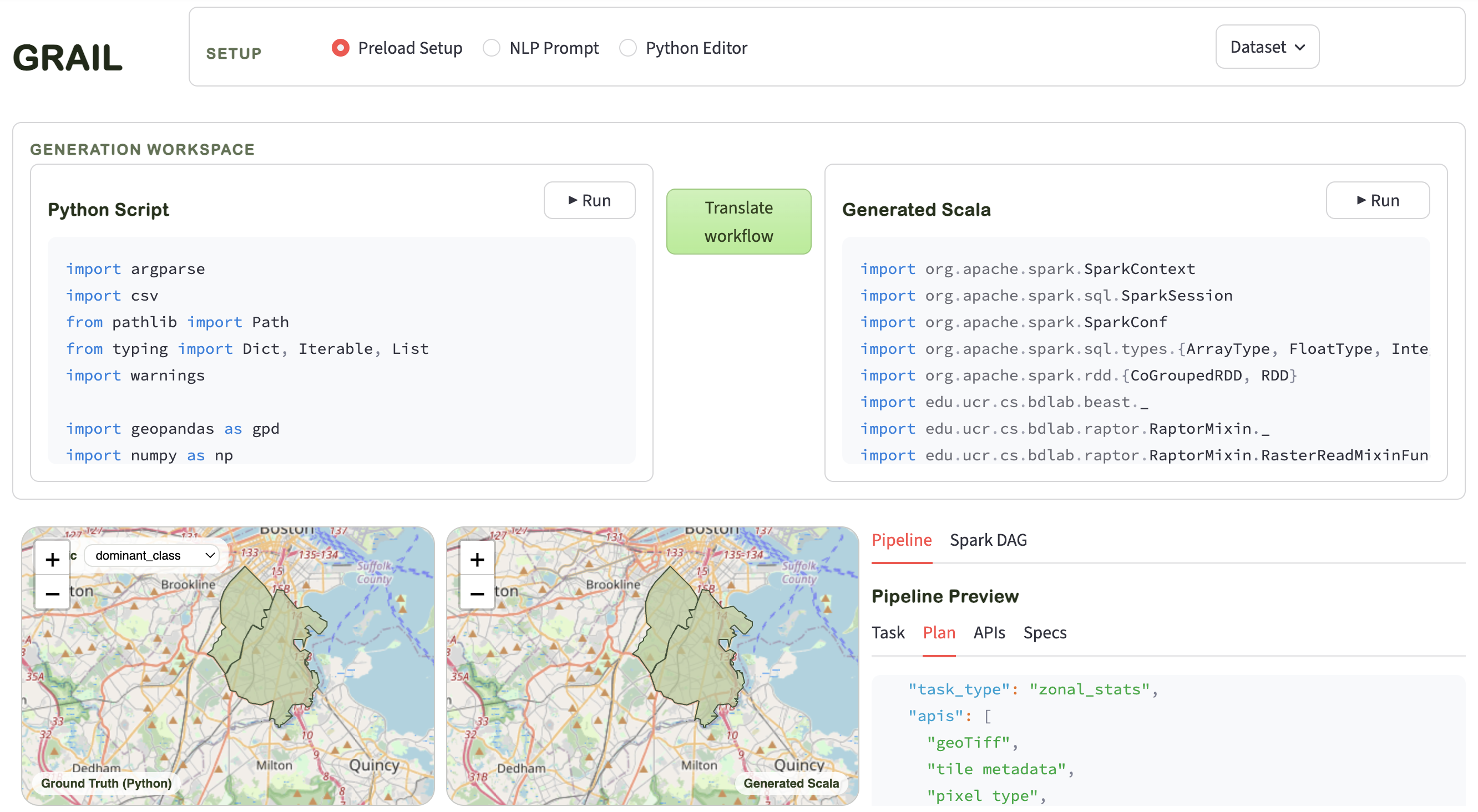}
    \caption{User Interface of GRAIL.}
    \label{fig:interface}
\end{figure*}
\section{Demo Scenarios}
\label{sec:demo}
\subsection{Boston Land Use: Translating Python to RDPro Scala}
In this demo scenario, we walk through a representative use case 
using our GRAIL interface. The user uploads a Python script that 
computes the percentage of each land use type across cities in 
the Boston area. 
As shown in Figure~\ref{fig:interface}, the interface presents 
the original Python script on the left alongside the generated Scala 
code on the right, produced by GRAIL translation pipeline. 
Bottom part of the figure shows the result comparison in a split map view, where 
users can inspect the Python and Scala outputs side by side to verify 
correctness. Users can also examine the pipeline steps selected by 
the agent and the Spark DAG generated during distributed execution. Our platform also supports plain-text translation and raw Python script editing using the setup panel at the top. Users also have the free choice of choosing a different dataset or using the pre-loaded one for experience.

\begin{figure}[t]
    \centering
    \includegraphics[width=0.85\columnwidth]{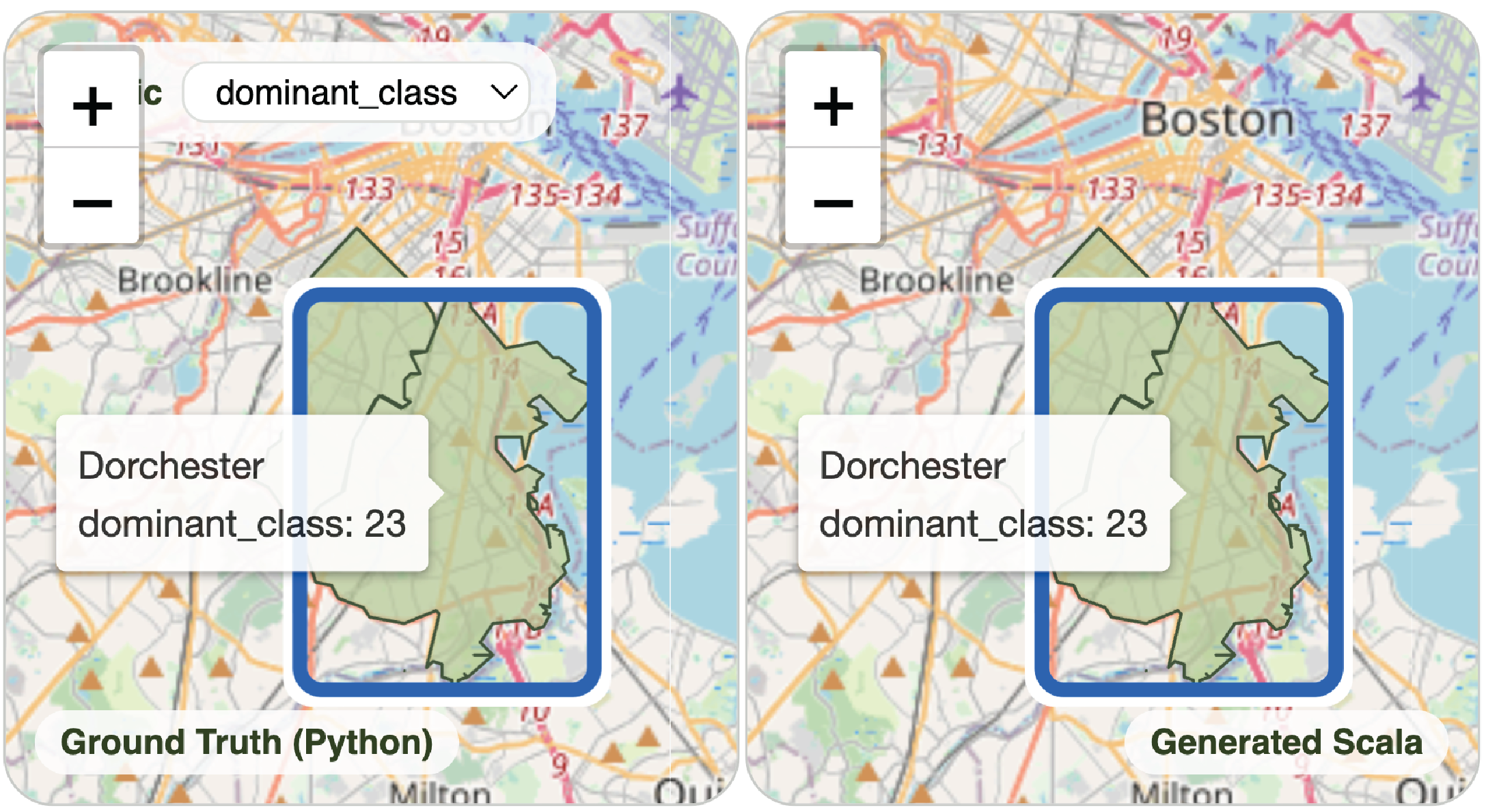}
    \caption{Comparison of dominant land use classification for Roxbury and Dorchester. 
Left: ground truth generated using Python. Right: result produced by LLM-generated Scala code.}
    \label{fig:nlcd}
\end{figure}

In this experiment, we use two cities, Roxbury and Dorchester, as sample study areas. The NLCD dataset is cropped with a spatial extent larger than the bounding boxes of these two cities to ensure full coverage during spatial operations. Figure~\ref{fig:nlcd} presents a split-view comparison, the left image shows the ground truth result produced using Python, while the right image shows the output generated by the GRAIL Scala code. Both results consistently indicate that the dominant land use class is label 23, corresponding to Developed, Medium Intensity.


\subsection{GRAIL Performance Across Different Dataset Sizes}
\begin{table}[t]
\centering
\caption{Vector dataset characteristics.}
\label{tab:datasets}
\begin{tabular}{lrrr}
\toprule
\textbf{Dataset} & \textbf{Polygons} & \textbf{Segments} & \textbf{Size} \\
\midrule
US counties      & 3,108   & 51,638     & 978\,KB \\
US states        & 49      & 165,186    & 2.6\,MB \\
World boundaries & 284     & 3,817,412  & 60\,MB  \\
US Census tracts & 74,133  & 38,467,094 & 632\,MB \\
\bottomrule
\end{tabular}
\end{table}
\begin{table}[t]
\centering
\caption{Zonal statistics runtime: Python vs.\ GRAIL on global Landsat Treecover (30\,m, 782\,GB).}
\label{tab:benchmark}
\begin{tabular}{lrrr}
\toprule
\textbf{Dataset} & \textbf{Python (s)} & \textbf{GRAIL (s)} & \textbf{Speedup} \\
\midrule
US counties      & 290   & 206   & 1.4$\times$ \\
US states        & 436   & 194   & 2.2$\times$ \\
World boundaries & 28,805 & 3,277 & 8.8$\times$ \\
US Census tracts & 1,001 & 210   & 4.7$\times$ \\
\bottomrule
\end{tabular}
\end{table}

We evaluate the scalability of GRAIL-generated code on a raster–vector join task across vector datasets of increasing size and polygon complexity, as shown in table~\ref{tab:datasets}, using Landsat Treecover as the raster input. During the demo, users can express a task such as \textit{``compute the average tree cover for each US census tract''}, and GRAIL automatically translates it into an executable Scala program. The Python baseline, implemented using Rasterio, applies a clip operation over each polygon bounding box for every overlapping raster tile. To ensure correctness across tile boundaries, overlapping tiles are merged into a virtual mosaic per polygon. 

Table~\ref{tab:benchmark} reports the results. Overall, GRAIL-generated code achieves better performance. At census-tract scale, Python performance degrades due to repeated tile reads, as each tile is accessed once per overlapping feature across thousands mask operations. In contrast, the technique used by GRAIL backend reads each tile exactly once and processes all overlapping features in a single streaming pass. This also explains why GRAIL shows similar performance trends across counties, states, and census tracts, since raster I/O dominates and is performed only once per tile.

The distributed execution improves efficiency by emitting compact intermediate results per partition and merging them locally, reducing network shuffles. For world-scale polygons, Python took up to 8 hours, while GRAIL-generated RDPro ran much faster by avoiding repeated raster reopening and masking, demonstrating strong scalability.
\subsection{Python Script vs. Natural Language as Translation Input}
In this demo, we compare three generation strategies driven by the same natural language task description. The first task is a raster only land-use summary workflow, while the second task is the same task that calculate land-use category percentages for Boston neighborhoods from NLCD raster data.
All three modes share the same section-by-section generation loop and validation stack, and the differ exists in how the task representation fed into loop. 

\textit{NL-Scala} takes the natural language description as input, reasons over the text to understand the workflow structure, and matches
the inferred task intent against the documented API before entering the
generation loop. Only the resulting structured task description is passed into section generation.

\textit{Python-NL-Scala} first generates an intermediate Python script from the natural language input. The LLM then uses that Python code to understand the task structure, in place of reasoning directly from text. The rest of the pipeline is identical to NL-Scala.

\textit{Python-Scala} also generates a Python reference script, but skips the structured analysis stage entirely. The raw Python source is passed directly into the generation loop with only coarse fallback analysis, and the model is instructed to translate rather than reason about the task. It serves as a baseline to isolate the value of structured task understanding.

  \begin{table}[h]
  \centering
  \caption{Benchmark Comparison: NL-Scala vs.\ Python-NL-Scala vs.\ Python-Scala Translation (5 runs each)}
  \label{tab:translation-benchmark}
  \begin{tabular}{llcc}
  \toprule
  \textbf{Task} & \textbf{Mode} & \textbf{Success Rate} & \textbf{Avg.\ Fix Iters} \\
  \midrule
  \multirow{3}{*}{\shortstack[l]{Raster-Only}}
  & NL-Scala          & 5/5 & 1.6 \\
  & Python-NL-Scala   & 4/5 & 2.6 \\
  & Python-Scala      & 0/5 & 5.0 \\
  \midrule
  \multirow{3}{*}{\shortstack[l]{Raster-Vector}}
  & NL-Scala          & 5/5 & 1.0 \\
  & Python-NL-Scala   & 4/5 & 2.4 \\
  & Python-Scala      & 0/5 & 5.8 \\
  \bottomrule
  \end{tabular}
  \end{table}

\paragraph{Raster-Only task.} \textit{``Calculate overall land-use category 
percentages for the NLCD raster and write a CSV summary. This is a 
raster-only workflow with no polygon overlay. Compute class counts and 
percentages for the whole raster and save a tabular CSV output.''}
\paragraph{Raster-Vector task.} \textit{``Calculate land-use category 
percentages and summary statistics for Boston neighborhoods from NLCD 
raster data. Produce tabular CSV 
outputs with per-neighborhood class percentages and a neighborhood-level 
dominant class summary.''}

We evaluate success rate and average fix iterations over 5 runs per condition, and the results are reported in Table~\ref{tab:translation-benchmark}.
NL-Scala achieves perfect success on both tasks with the fewest fix iterations. Python-NL-Scala is competitive but requires more repair. This mode still benefits from structured analysis and section planning, but its reasoning is mediated through a generated Python reference. As
a result, the program introduce more ambiguities that
can affect the downstream understanding. Python-Scala
fails entirely on both tasks, confirming that raw translation without structured task reasoning is insufficient.

These results show that a Python intermediate does not improve
performance on its own. Natural language is a more direct carrier of task intent, and its value is most apparent when no intermediate artifact obscures the reasoning process.




\bibliographystyle{ACM-Reference-Format}
\bibliography{sample}

\end{document}